\documentclass[conference]{IEEEtran}

\IEEEoverridecommandlockouts                              

\usepackage{graphics} 
\usepackage{subfigure}
\usepackage{epsfig} 
\usepackage{mathptmx} 
\usepackage{times} 
\usepackage{amsmath} 
\usepackage{amssymb}  
\usepackage{float}
\usepackage{booktabs}
\usepackage{graphicx}
\usepackage{cite}
\usepackage{blindtext}
\usepackage[linesnumbered, ruled]{algorithm2e}
\usepackage{threeparttable}
\usepackage{enumitem}
\usepackage[backref]{hyperref}
\usepackage[capitalise]{cleveref}
\makeatletter
\let\NAT@parse\undefined
\makeatother

\title{\LARGE \bf
PALF: Pre-Annotation and Camera-LiDAR Late Fusion for the Easy Annotation of Point Clouds
}

\author{Yucheng Zhang$^{1,*}$, 
Masaki Fukuda$^{2}$,
Yasunori Ishii$^{2}$, 
Kyoko Ohshima$^{1}$, and
Takayoshi Yamashita$^{3}$
\thanks{}
\thanks{*Corresponding author: Yucheng Zhang(zhang.yucheng@jp.panasonic.com).}
\thanks{$^{1}$R\&D Division, Panasonic Automotive Systems Co., Ltd,
Yokohama City, Japan}
\thanks{$^{2}$Technology Division, Panasonic Holdings Co., Ltd, Osaka City, Japan}
\thanks{$^{3}$Department of Computer Science, Chubu University, Kasugai City, Japan}
}
\begin{document}
\maketitle

\begin{abstract}
3D object detection has become indispensable in the field of autonomous driving. To date, gratifying breakthroughs have been recorded in 3D object detection research, attributed to deep learning. However, deep learning algorithms are data-driven and require large amounts of annotated point cloud data for training and evaluation. Unlike 2D image labels, annotating point cloud data is difficult due to the limitations of sparsity, irregularity, and low resolution, which requires more manual work, and the annotation efficiency is much lower than 2D image. Therefore, we propose an annotation algorithm for point cloud data, which is pre-annotation and camera–LiDAR late fusion algorithm to easily and accurately annotate. The contributions of this study are as follows. We propose (1) a pre-annotation algorithm that employs 3D object detection and auto fitting for the easy annotation of point clouds, (2) a camera–LiDAR late fusion algorithm using 2D and 3D results for easily error checking, which helps annotators easily identify missing objects, and (3) a point cloud annotation evaluation pipeline to evaluate our experiments. The experimental results show that the proposed algorithm improves the annotating speed by 6.5 times and the annotation quality in terms of the 3D Intersection over Union and precision by 8.2 points and 5.6 points, respectively; additionally, the miss rate is reduced by 31.9 points.
\end{abstract}

\section{INTRODUCTION}

Light detection and range (LiDAR) sensors can be employed to obtain accurate distance information (i.e., distance error $<$ 2cm), which is essential for ensuring the safety of autonomous driving systems, path planning, and decision making. Correspondingly, LiDAR sensors have gained popularity in the field of autonomous driving.

As deep learning has achieved impressive results in the imaging field in recent years, the adoption of deep learning to process the point cloud data obtained by LiDAR sensors for perception tasks has also become a research hotspot. However, deep learning is a well-known data-driven algorithm that is always data-hungry. Generally, there are currently some public datasets, such as KITTI \cite{c2}, ApolloScape \cite{c3}, nuScenes \cite{c4}, Waymo \cite{c5}, and Pandaset \cite{c6}, which are open for researchers or enterprises to use. However, the application of the models trained on these public datasets to perform certain tasks is limited by the 3D object detection domain shift problem, which decreases the models’ accuracy. We need to retrain the 3D object detection model with more local data (LiDAR sensor type, weather, and geographic location differences often lead to a significant reduction in the accuracy of the same 3D object detection model\cite{c7}\cite{c8}).
Therefore, considerably more annotated point cloud data are still required, and the development of an algorithm for point cloud annotation that is both easy and accurate is crucial.

Although many research papers on image annotation have been published \cite{c9}\cite{c10}, the studies focusing on point cloud annotation are few. Different from labeling images, annotating point cloud data is tedious and challenging. Conventionally, when objects in an image are labeled, only a 2D box needs to be drawn to complete most operations. In comparison, the annotation of point cloud data is more complicated. As point cloud data have six degrees of freedom, the sparseness and irregularity of the point cloud data make the annotation process very time-consuming and burdensome. In addition, mistakes tend to be made in manual annotation. For instance, in \cite{c11}, \cite{c12}, and \cite{c13}, it is mentioned that the KITTI 3D detection dataset \cite{c2} provides the wrong ground truth (i.e., the position offset of the 3D box).

\begin{figure*}[t]
\centering
\subfigure[Low-resolution] 
{
\includegraphics[width=6cm]{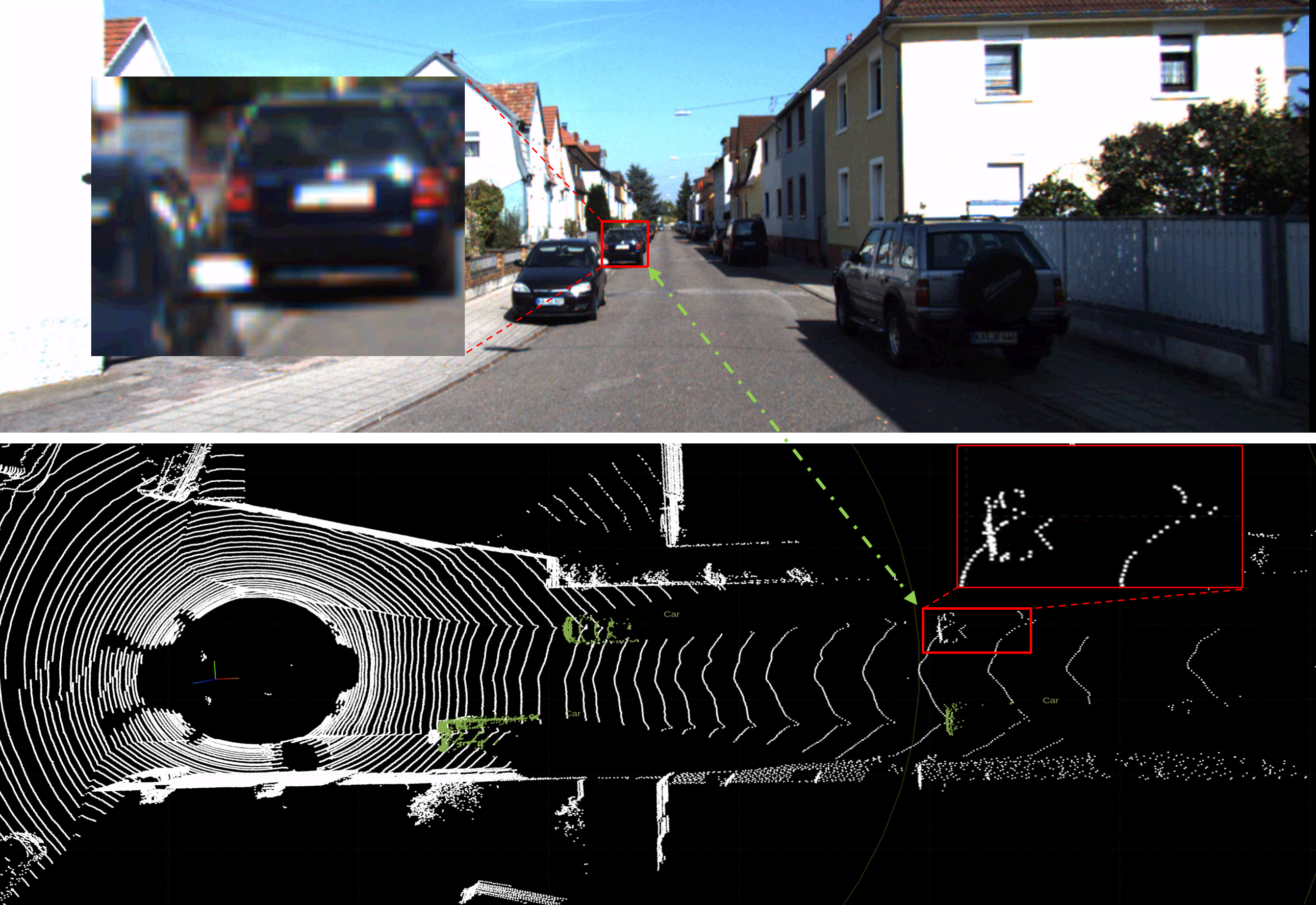}} 
\subfigure[Complex operation]{
\includegraphics[width=5cm]{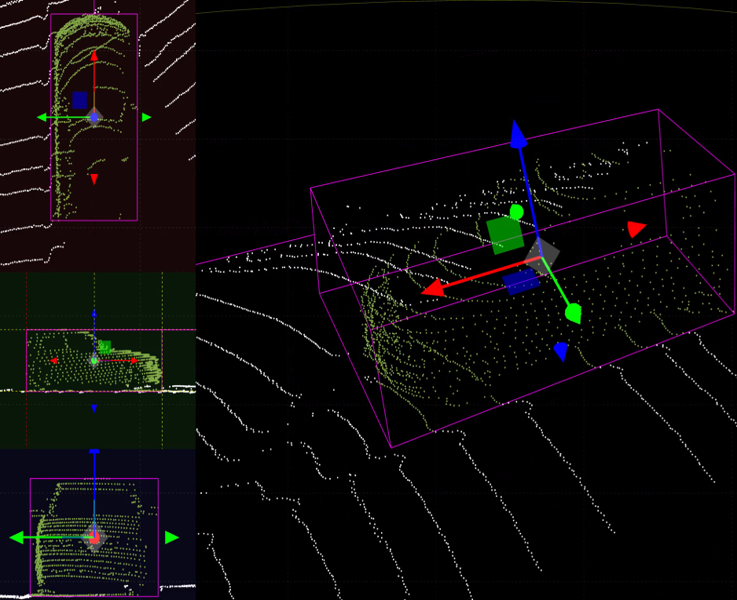}}
\subfigure[Repeatability]{
\includegraphics[width=4.4cm]{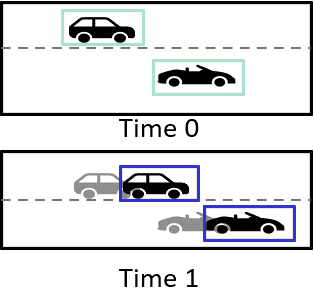}}
\caption{Challenges in point cloud annotation. (a) The sparsity of the point cloud makes it difficult for annotators to identify objects. (b) The annotator must fine-tune the 3D box with six degrees of freedom, which is a very tedious and exhausting operation. (c) Objects are annotated at times 0–1, and even if the same object has been annotated at time 0, it must be annotated again at time 1.} 
\label{1}  
\end{figure*}

There are three main challenges in annotating point cloud data:  \textbf{1) Low-resolution:} In Fig.~\ref{1}(a), we can quickly identify the object in the 2D image. However, in the point cloud case, the sparsity problem makes it difficult for annotators to identify objects clearly. 
\textbf{2) Complex operation:} As shown in Fig.~\ref{1}(b), since the point cloud comprises 3D data, it is difficult to confirm whether the 3D box correctly and completely contains the object. Thus, it is necessary to adjust the 3D box by six degrees of freedom finely, although this process is quite tedious and exhausting. 
\textbf{3) Repeatability:} As shown in Fig.~\ref{1}(c), point cloud data are generally obtained in successive frames (~\cite{c5} and ~\cite{c6}); therefore, in most cases, the same objects in different frames will be repeatedly annotated.

If these challenges remain unaddressed, it will be difficult to efficiently annotate point cloud data, which will significantly limit the progress in 3D object detection. To address these challenges, some researchers have proposed semi-automatic and manual point cloud annotation tools, as well as algorithms to facilitate annotation \cite{c11}\cite{c12}\cite{c13}\cite{c14}\cite{c15}\cite{c16}\cite{c17}. However, these annotation tools only partially address the challenges, and the algorithms tend to fail in many cases. Additionally, annotators must identify each object very carefully, which is strenuous and quite challenging.

\begin{figure}[]
\includegraphics[width=1\linewidth]{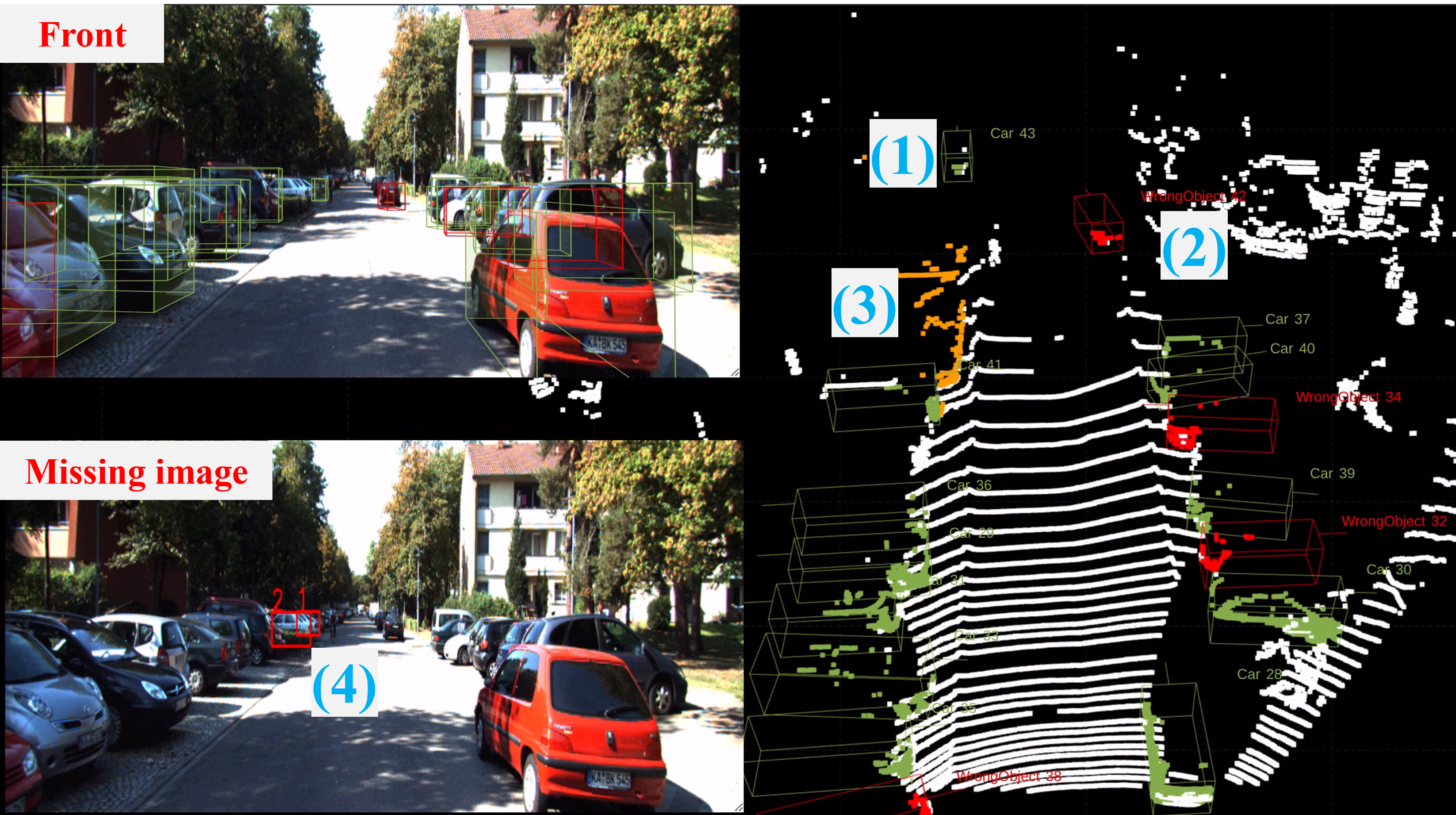} 
\caption{The main user interface of the PALF algorithm. (1) The green boxes indicate the pre-annotation result. (2) The red boxes indicate the camera unmatched result or that the Intersection over Union (IoU) is lower than the threshold. (3) The orange markings in the point cloud represent the missed-object regions from the image back-projection. (4) The missed-object regions in the image mentioned in (3).
} 
\label{2}
\end{figure}

Therefore, as a solution, we proposed the pre-annotation and camera–LiDAR late fusion algorithm, as shown in Fig.~\ref{2}. Our algorithm builds on the work of \cite{c11}. Compared with those in \cite{c11}, our algorithm can increase the annotation speed by 6.5 times, 3D Intersection over Union (IoU) by 8.2 points, precision by 5.6 points, and recall by 31.9 points.

Our work contributes the following:

\textbf{1) A pre-annotation algorithm.} By employing an off-the-shelf 3D object detection model and a fitting algorithm to generate the initial 3D box, the required steps in finding objects in point clouds and the manual operation time when using the rule-based method can be reduced.

\textbf{2) A camera–LiDAR fusion algorithm.} By fusing the results of 2D object detection and 3D object detection, accuracy improvements can be obtained with 2D object detection than with 3D object detection to help annotators more efficiently check for errors and missed objects.

\textbf{3) A pipeline for evaluating the point cloud annotation algorithm.} Our proposed algorithm is tested on the KITTI dataset, and a comparative study with the baseline in \cite{c11} reveals that our algorithm significantly improves the annotation efficiency and quality. Additionally, each of our distinct features is validated through detailed ablation studies.

The remaining parts of this paper are organized as follows. In Section II, we review the related works on point cloud annotation. In Section III, we describe the system architecture and main functions of our algorithms. In Section IV, we discuss the ablation study evaluation results of our research and provide a summary of related discussions. In Section V, we conclude the paper.

\section{RELATED WORK}
In this section, we provide a comprehensive analysis of the popular point cloud annotation tools. As shown in TABLE ~\ref{I}, eight tools were compared based on the four limiting features: \textbf{low-resolution}, \textbf{complex operation}, \textbf{repeatability} and \textbf{github stars}, and SUSTechPOINTS was selected as the baseline.

\begin{table}[t]
  \caption{A comparative study of point cloud annotation tools and the selection of the best baseline SUSTechPOINTS. \textbf{Multi-View} indicates if the annotation tool performs annotation from multiple perspectives. \textbf{Fusion} indicates if there is a camera auxiliary annotation function. The \textbf{Github Stars} refers to conventionally, the higher the number of stars, the easier it is to use and install the tool.}
  \label{I}
  \includegraphics[width=1\linewidth]{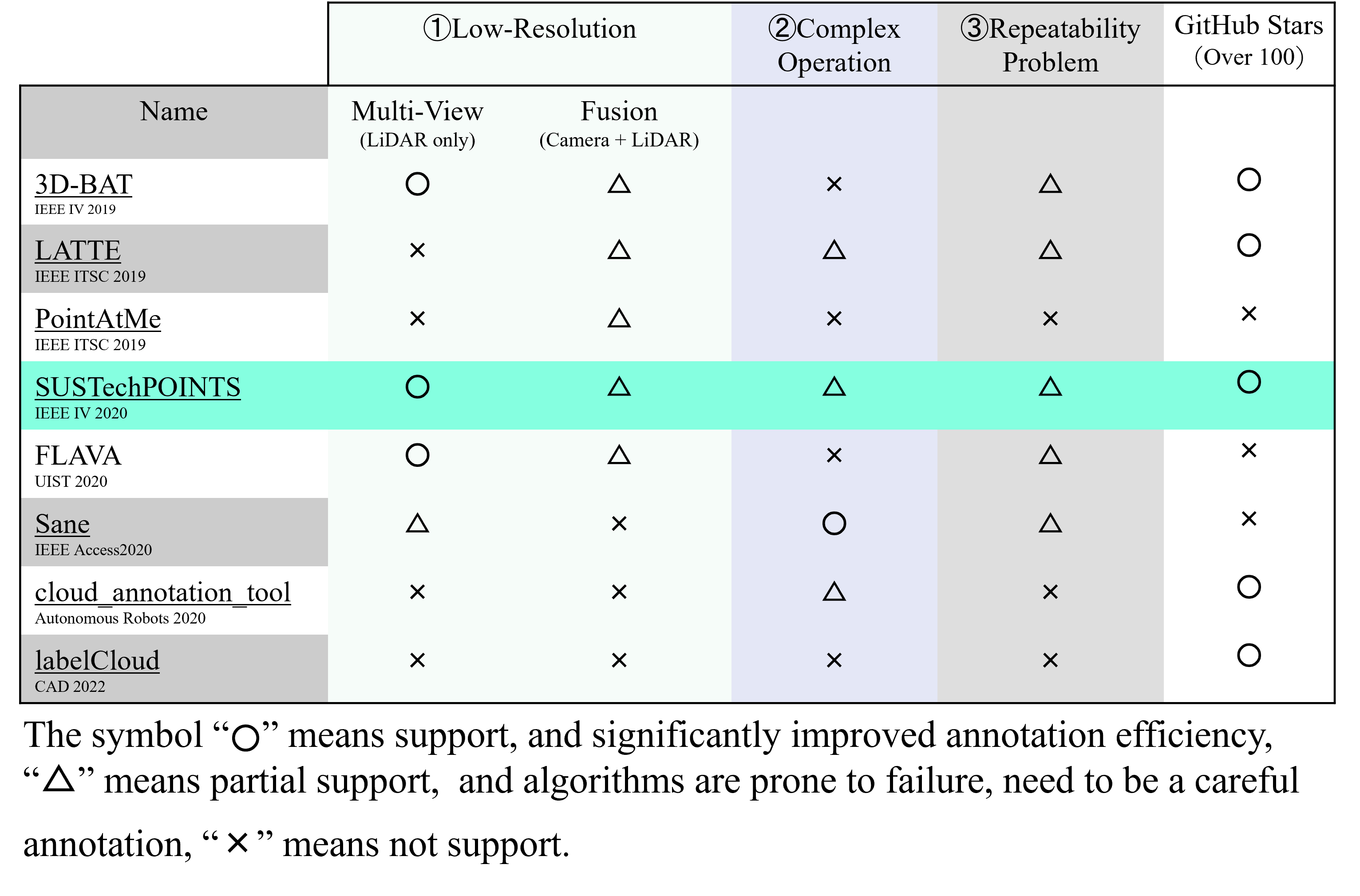} 
\end{table}

 \textbf{1) Low-resolution problem:} tools such as those introduced in \cite{c11}\cite{c12}\cite{c14} provide solutions for bird-eye view, side view, and front view to enable annotators to efficiently determine the annotation accuracy of objects. Additionally, the information from the camera is employed to reduce the time required to find the target object and assist the annotators. However, all these tools ignore the fact that identifying objects to annotate from point clouds is tedious. Due to the sparsity and irregularity of the point cloud, it is difficult to easily find the object we intend to annotate, even using a camera.

\textbf{2) The complex operation problem:} \cite{c15} proposed the use of DBSCAN \cite{c19}, and \cite{c11} employed a fitting algorithm to perform one-click annotation of the point cloud. 
However, although these algorithms significantly reduce the number of required annotation operations, they are rule-based and tend to fail. For instance, examining the fitting algorithm in \cite{c11}, the final generated box fails if a size area slightly larger or smaller than the object is employed. Additionally, the operation problem still exists because it is necessary to find each object in the point cloud and manually label it separately, which is very time-consuming and labor-intensive.

\addtolength{\topmargin}{0.05in}

\textbf{3) Repeatability problem:}~\cite{c12} proposed the greedy-search algorithm, \cite{c11} and \cite{c13} proposed the transfer algorithm, and \cite{c15} employed the Kalman filter~\cite{c23} to track the same object from times 0 to 1. Their core idea was to transfer the annotated data of the first frame (or the previous frame) to the next frame to solve the repeatability problem. However, all these algorithms are rule-based; consequently, they tend to fail during the transfer process, as is the case in \cite{c11}, \cite{c13}, and \cite{c15}. Further, after using the transfer algorithm, the center point of the 3D box will be offset. In addition, the algorithm generally needs to annotate the object of several frames in advance to achieve the optimal effect. Furthermore, annotators are often overburdened since offset center points and advanced markings require human operation.

\section{SYSTEM ARCHITECTURE AND ALGORITHM}

In this section, we provide details of the two algorithms: \textbf{pre-annotation} and \textbf{camera–LiDAR late fusion}.

\subsection{Pre-Annotation} 

\begin{figure*}[thpb]
\includegraphics[width=1\linewidth]{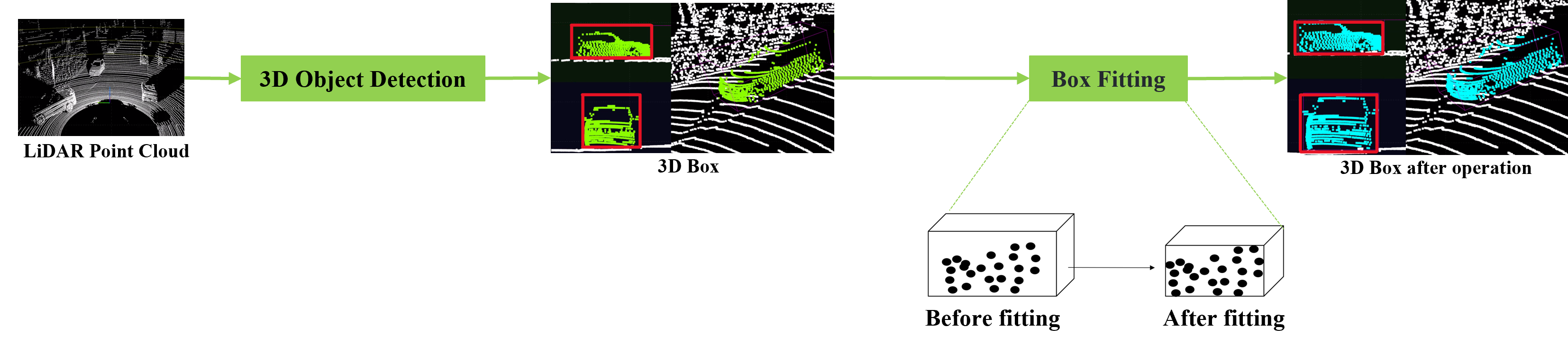} 
\caption{The pipeline of Algorithm 1, Pre-Annotation. We first input point cloud into the off-the-shelf 3D object detection model. After obtaining the 3D box, we fit the box by the fitting algorithm. Finally, we generate the pre-annotated result, which is best viewed in color.} 
\label{3}
\end{figure*}

\textbf{Motivation:} Recently, deep learning has achieved surprising accuracy in the object detection field. Although the results directly outputted by 3D object detection systems cannot be considered as ground truth, they are still beneficial as they help the annotator find the target object, which is equivalent to providing many pre-annotated 3D boxes. The annotator only needs to determine if the algorithm-generated box is accurate. 

\SetKwRepeat{Do}{do}{while}%
\begin{algorithm}[t]
\SetKwInput{KwData}{Input}
\SetKwInput{KwResult}{Output}
\KwData{Point cloud $P \in R^{n\times3}$, Fitting threshold $\mathrm{Point}^{threshold}$}
  \KwResult{Box after operation $B^{\prime}$($p, s, r$), where $p, s, r$ $\in$ $ R^{n\times3}$ correspond to the position, scale, and rotation. }
  {input $P$ into the off-the-shelf 3D object detection model to obtain the initial box $B$($p, s, r$)}\;
  \For{object in all objects}{
    \If{points of object $>$ $\rm{Point}^{threshold}$}
    {
    fitting object box with fitting algorithm \cite{c11}, get $B^{\prime}$($p, s, r$)\;
    }
  }
  \Return $B^{\prime}$\;
  \caption{Pre-Annotation Algorithm}
  \label{Algorithm 1}
\end{algorithm}

We propose a pre-annotation algorithm using 3D object detection to improve annotation convenience, as shown in Fig.~\ref{3}. 
First, we input the point cloud into the 3D object detection model, which outputs a 3D box with seven parameters (x, y, z, length, width, height, and yaw angle). Notably, the outputted 3D box is always slightly larger than is required or has shortcomings, such as inaccurate center points. Therefore, to obtain an accurate 3D box, we use the fitting algorithm implemented in \cite{c11} for matching. Since the fitting will fail if only a few points are available, we only employ the fitting algorithm for the box whose number of points exceeds the threshold. See Algorithm 1 for details.

\subsection{Camera-LiDAR Late Fusion}
\SetKwRepeat{Do}{do}{while}%
\begin{algorithm}[t]
\SetKwInput{KwData}{Input}
\SetKwInput{KwResult}{Output}
  \KwData{Point cloud $P \in R^{n\times3}$, Calibrated image, Initial $\mathrm{box}_{3D}^{\mathrm{LiDAR}}(p,s,r)$,  $\mathrm{IoU}_{2D}^{\mathrm{threshold}}$ }
  \KwResult{Box after operation $B^{\prime}(p, s, r)$, Point cloud after operation $P^{\prime} \in R^{n\times3}$ }
  {$\mathrm{box}_{3D}^{\mathrm{LiDAR}}(p,s,r)$ projection into image $\mathrm{box}_{2D}^{\mathrm{LiDAR}}(x,y)$}\;
  {calibrated image inputted into the off-the-shelf 2D object detection model affords $\mathrm{box}_{2D}^{\mathrm{image}}$}\;
      {calculate the Euclidean distance, $\mathrm{distance}_{center}$ between center of $\mathrm{box}_{2D}^{\mathrm{LiDAR}}$ and $\mathrm{box}_{2D}^{\mathrm{image}}$}\;
  {matched $\leftarrow \mathrm{Hungarian}\ \mathrm{Algorithm}(\mathrm{distance}_{center})$}\;
  
\For{object in all objects}{
 \If{$\mathrm{box}_{2D}^{\mathrm{LiDAR}}$ and $\mathrm{box}_{2D}^{\mathrm{image}}$ matched}
   {{$\mathrm{IoU} \leftarrow \mathrm{IoU}(\mathrm{box}_{2D}^{\mathrm{LiDAR}},\mathrm{box}_{2D}^{\mathrm{image}})$}\;
       \If{$\mathrm{IoU} < \mathrm{IoU}_{2D}^{\mathrm{threshold}}$}{
       $\mathrm{wrong\ list} \leftarrow \mathrm{Index\ of\ IoU}$
       }

}
{$\mathrm{wrong\ list} \leftarrow \mathrm{LiDAR\ unmatched\ index}$}\;
{$\mathrm{missed\ list} \leftarrow \mathrm{Camera\ unmatched\ index}$}\;
}
\If{wrong list not null}{
highlight the $\mathrm{box}_{3D}^{\mathrm{LiDAR}}$ according to the wrong list, get $B^{\prime}(p,s,r)$\;
}
\If{missed list not null}{
{point cloud projection into the image, remove points outside the image boundaries, get the point cloud projected image (2D)}\;
{the box area in the missed list is back-projected to the point cloud and highlighted, get $P^{\prime}$}\;
}
\Return $B^{\prime}, P^{\prime}$\;
\caption{Camera-LiDAR Late Fusion Algorithm}
\label{Algorithm 2}
\end{algorithm}

\textbf{Motivation:} High-precision 2D object detection has been achieved in the object detection field, with generally higher accuracy than that for 3D object detection. Thus, we integrate the 2D object detection results of the image with the 3D object detection results of the point cloud and use the high-precision model of the 2D object detection to facilitate the point cloud annotation. This will enable the annotator to easily check for errors and find missed objects. 

\subsubsection{\textbf{Late fusion part}}

\begin{figure*}[thpb]
\includegraphics[width=1\linewidth]{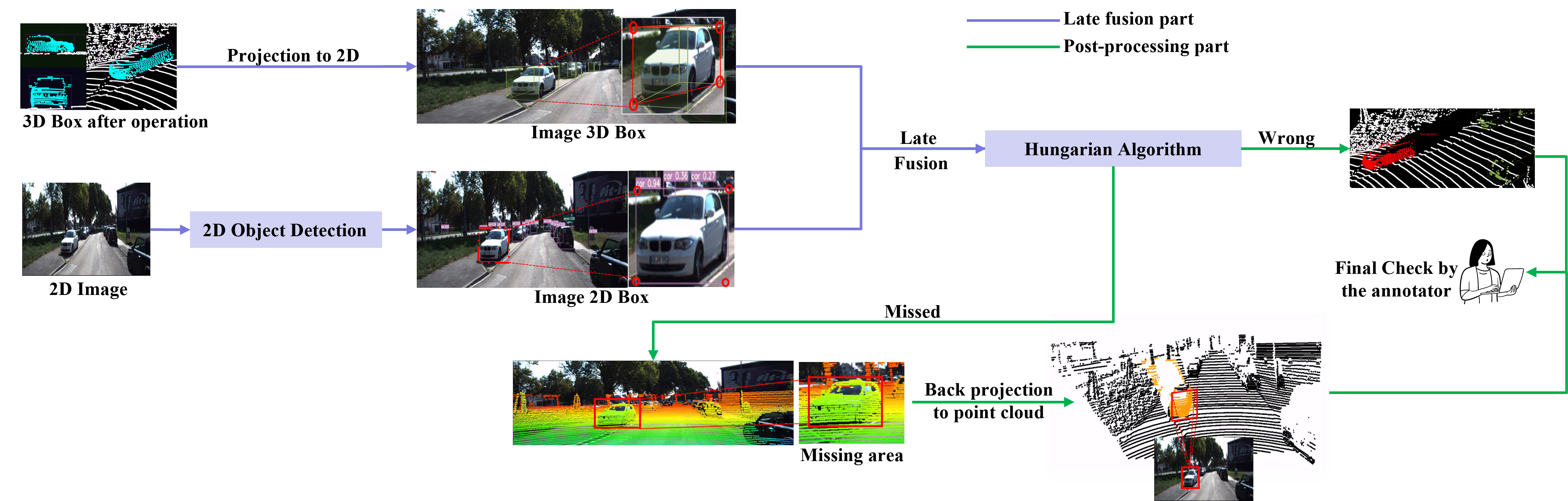} 
\caption{The pipeline of the camera–LiDAR late fusion algorithm. First, the 3D box is projected to the image plane. Thereafter, the calibrated image is inputted into the off-the-shelf 2D object detection model. After the late fusion, the wrong and missed results are obtained, and the annotator is notified after post-processing.} 
\label{4}
\end{figure*}

As shown in the part represented by the purple line in Fig.~\ref{4}, we back-project the 3D box obtained in the pre-annotation stage into the image and use the camera calibration formula mentioned in \cite{c2} to generate the 3D box of the image. Thereafter, we input the image calibrated with the point cloud into the 2D object detection model to generate 2D boxes. Since we have 3D boxes in the image plane from the pre-annotation stage and 2D boxes from the 2D object detection, the fusion of these two bounding boxes can result in an assignment problem (i.e., the mismatch between the 2D results and 3D results).

The Hungarian algorithm \cite{c21} is well suited to solving such assignment problems. Since the numbers of 2D and 3D results do not always match, we slightly modify the algorithm from \cite{c21}. If the matrix is not a square, it will automatically fill 0 to any row/column. Regarding the cost matrix of the Hungarian algorithm, we take the center point formed from the 3D box in the image plane and the center point of the 2D box, calculate the Euclidean distance, determine the distance between the 2D box and the 3D box, and output the 3D box as well as the matching result of the 2D box, as shown in Algorithm 2 lines 3–4.

\subsubsection{\textbf{Post-processing part}}



Our post-process is as shown in Algorithm 2, lines 5–14.
\textbf{(i) Matched. }
To eliminate some edge cases, such as situations where people are close to the car, we perform 2D IoU calculation of the quadrilateral formed by the outermost points of the 3D box in the image plane and the quadrilateral formed by the 2D box. If the IoU exceeds a certain threshold, it is regarded as ground truth. If the IoU threshold is not exceeded, the 3D box is put into the wrong list. 
\textbf{(ii) Unmatched 3D box.} Since the accuracy of the 2D object detection model is comparatively good, we put the 3D box into the wrong list.
\textbf{(iii) Unmatched 2D box.} Similarly, we treat the 2D box as a missed-object and put it in the missed list.
\textbf{(iv) Unmatched.} The input data may be incorrect.

According to the wrong list and missed list, the boxes are processed separately, as shown in Algorithm 2 lines 15–21. Regarding the wrong list, as shown in Fig.~\ref{4}, post-processing occurs after the green arrow labeled wrong, and the point cloud in the 3D box is highlighted in red. Compared with the surrounding, correct green 3D box, the red 3D box can easily be found, which is convenient for the annotator. Regarding the missed list, as shown in Fig.~\ref{4}, post-processing occurs after the green arrow labeled missed. First, we project the point cloud into the image plane by camera calibration formula \cite{c2}. Thereafter, we generate the missed area according to the missed list obtained from the late fusion. Finally, the missed areas are back-projected to the point cloud and are highlighted in orange. Simultaneously, we display the missed box in the image so that the annotator can find the missed area easily.

In conclusion, through our proposed PALF algorithm, pre-annotated 3D boxes are generated to facilitate convenient annotation, which reduces the time required to find the objects and manually annotate them individually in low resolution and alleviates the problem of the cumbersome manual-labeling operation. Further, regarding the repeatability problem, there is no need to annotate in advance to improve the effectiveness of the algorithm. In addition, by fusing the 2D and 3D results and using the results of high-precision 2D object detection to provide automatic error checking, annotators can quickly find errors and missed objects, which will prevent the manual, burdensome checking process in low resolution and complex operation problems. Additionally, the algorithm reduces the time required to find objects for annotation, improving the annotation efficiency and quality.
\section{EXPERIMENTS}

We evaluated the effectiveness of the proposed algorithm in terms of the annotation quality and efficiency.
\subsection{Experimental setup}

To evaluate our algorithm, we refer to KITTI dataset frames 000010, 000011, 000016, and 000017, which were employed in \cite{c11} and \cite{c16} experiments. However, in \cite{c11} and \cite{c16}, overly few objects were selected in the four frames of KITTI (i.e., eight cars, two cars, six cars, and one car, totaling only 17 cars). Real-world scenes are conventionally very complex, with many vehicles. To ensure consistency with real-world labeled scenarios, we added four other frames: 007216, 007266, 007458, and 007479, with 37 cars from the KITTI dataset. Finally, we experiment with the data of eight frames from KITTI.

As mentioned in \cite{c11}\cite{c12}\cite{c15}\cite{c16}, the KITTI ground truth is inaccurate. Therefore, we asked an annotation expert to carefully annotate the experimental data as the ground truth data. We selected three annotators for the annotation. Before running the experiment, we trained each annotator for approximately 1 hour to familiarize them with the annotation tool. Furthermore, they were asked to annotate only objects that they were sure of (it is difficult to annotate distant objects because of the nature of the point cloud).

To prove that our algorithm can significantly improve the annotation efficiency and quality, we initially asked the three annotators to employ all the functions in \cite{c11} to label the eight frames and obtain the baseline annotation time and quality. Subsequently, after 10 days, the annotators were instructed to use our proposed algorithm to annotate the same data.

For the 3D object detection model, we used voxel-RCNN \cite{c1} to pre-train the model on the Pandaset dataset. For the 2D object detection, we used YOLOv7 \cite{c20}.

\subsection{Metrics}

For the annotation efficiency, we considered the time required to evaluate an instance. For the annotation quality, we considered the 3D IoU, precision, and recall. In addition, we considered the miss rate(also called false negative rate) and number of objects for the evaluation.

The 3D IoU algorithm is provided by PyTorch3D\cite{c22}.

\subsection{Results}
As shown in TABLE~\ref{II}, compared with our experts’ data, the 3D IoU of the KITTI data is 74.3\%, precision is 79.2\%, and the miss rate is 23.6\% of the object.

Compared with the baseline \cite{c11}, we found that after adding the PALF algorithm, both the efficiency and quality were improved. For the annotation efficiency, the total time was reduced by 70.0\%, and the time/object was reduced by 84.7\%; for the annotation quality, the 3D IoU increased by 8.2 points, precision has increased by 5.6 point, recall has increased by 31.9 point; and the number of objects increased by a factor of 1.96.

\subsection{Ablation Study}
We conducted an ablation study in three parts, and the results are shown in TABLE~\ref{II}:
\begin{enumerate}
\item Pre-annotation (PA)
\item Pre-annotation + wrong check (PAWC)
\item Pre-annotation + wrong check + missing check (PALF)
\end{enumerate}

\begin{table*}[t]
\caption {Results of the PALF algorithm} \label{tab:title}
\resizebox{\linewidth}{!}{ 
\begin{tabular}{l|c|c|c|c|c|c|c}
\hline
\multicolumn{1}{c|}{Method} & \begin{tabular}[c]{@{}c@{}}Time\\ (s)\end{tabular} & \begin{tabular}[c]{@{}c@{}}3D IoU\\ (\%)\end{tabular} & \begin{tabular}[c]{@{}c@{}}Precision\\ (\%)\end{tabular} & \begin{tabular}[c]{@{}c@{}}Recall\\ (\%)\end{tabular} & \begin{tabular}[c]{@{}c@{}}Miss Rate\\ (\%)\end{tabular} & \begin{tabular}[c]{@{}c@{}}Number of  objects\\ (vehicle)\end{tabular} & \begin{tabular}[c]{@{}c@{}}Time per object\\ (s)\end{tabular} \\ \hline
\multicolumn{1}{c|}{KITTI} & - & 74.3 & 79.2 & 76.4 & 23.6 & 53 & - \\ \hline
SUSTechPOINTS Baseline & 3334 & 75.9 & 78.3 & 31.6 & 68.4 & 24.3 & 137.2 \\ \hline
1) PA & 1597 & 84 & \textbf{88.6} & 65.7 & 34.3 & 44.7 & 35.7 \\
2) PAWC & 1321 & 83.8 & 86.1 & \textbf{68.2} & \textbf{31.8} & 47.7 & 27.7 \\
3) PALF (Full features) & \textbf{1000}\ & \textbf{84.1} & 83.9 & 63.5 & 36.5 & \textbf{47.7} & \textbf{21} \\ \hline
\end{tabular}
}
\label{II}
\end{table*}

Compared with the baseline values, after adding 1), 2), and 3), the total annotation time was reduced by 52.1\%, 60.4\%, and 70.0\%, and the time/object was reduced by 74.0\%, 79.8\%, 84.7\%, respectively. For the annotation quality, the 3D IoU increased by 8.1 points, 7.9 points, and 8.2 points; the precision increased by 10.3 points, 7.8 points, and 5.6 points; and the recall increased by 34.1 points, 36.6 points, 31.9 points, respectively. We found that regardless of which features are added, the annotation speed and quality are significantly improved.

\begin{figure*}[t]
\includegraphics[width=1\linewidth]{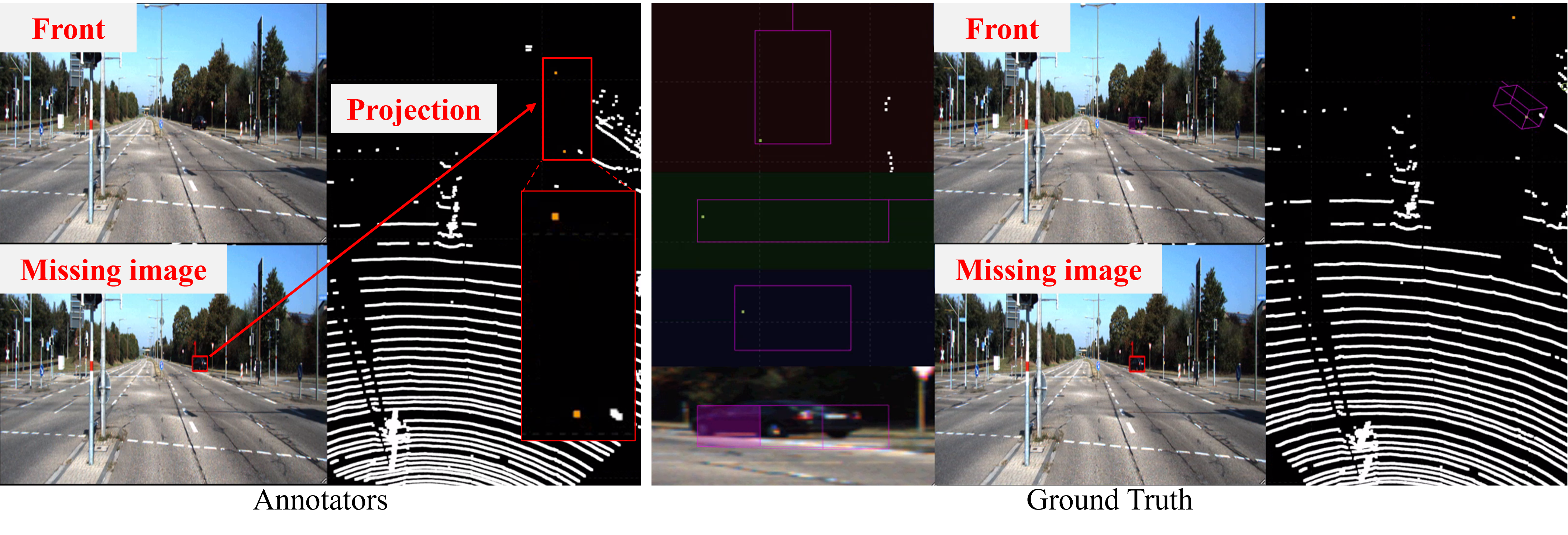} 
\caption{The analysis miss rate did not decrease. As shown in the annotators part of the figure, the 2D detection model provided missed objects; however, only two orange markings can be observed in the point cloud, annotators are not confident in annotating this missed object. The ground truth data on the right side of the picture comes from experts’ data.
}
\label{5}
\end{figure*}

Notably, after adding the full features, the annotation time/efficiency and 3D IoU improved; however, the precision and recall were not as high as those obtained using the pre-annotation algorithm alone. In particular, the recall rate did not improved considerably. As shown in Fig.~\ref{5}, after analyzing the results, we found that although the 2D object detection model afforded many hints on the missing objects, not many are projected into the point cloud. This is because in a relatively distant place, the point cloud is rarely or 0 point. These missing objects still pose a traditional complex problem in point cloud research.

\section{CONCLUSIONS}

In this work, we introduced the PALF algorithm, which improves point cloud annotation. The results of ablation studies proved the effectiveness of our algorithm, confirming that the proposed algorithm significantly improves the annotation efficiency and quality, as well as the ease of annotation. In future work, the information on the time axis can be exploited to perform box backpropagation on sparse point clouds.

\bibliographystyle{IEEEtran}
\bibliography{reference}

\begin{thebibliography}{10}
\providecommand{\url}[1]{#1}
\csname url@samestyle\endcsname
\providecommand{\newblock}{\relax}
\providecommand{\bibinfo}[2]{#2}
\providecommand{\BIBentrySTDinterwordspacing}{\spaceskip=0pt\relax}
\providecommand{\BIBentryALTinterwordstretchfactor}{4}
\providecommand{\BIBentryALTinterwordspacing}{\spaceskip=\fontdimen2\font plus
\BIBentryALTinterwordstretchfactor\fontdimen3\font minus
  \fontdimen4\font\relax}
\providecommand{\BIBforeignlanguage}[2]{{%
\expandafter\ifx\csname l@#1\endcsname\relax
\typeout{** WARNING: IEEEtran.bst: No hyphenation pattern has been}%
\typeout{** loaded for the language `#1'. Using the pattern for}%
\typeout{** the default language instead.}%
\else
\language=\csname l@#1\endcsname
\fi
#2}}
\providecommand{\BIBdecl}{\relax}
\BIBdecl

\bibitem{c2}
A.~Geiger, P.~Lenz, C.~Stiller, and R.~Urtasun, ``Vision meets robotics: The
  kitti dataset,'' \emph{The International Journal of Robotics Research},
  vol.~32, no.~11, pp. 1231--1237, 2013.

\bibitem{c3}
X.~Huang, P.~Wang, X.~Cheng, D.~Zhou, Q.~Geng, and R.~Yang, ``The apolloscape
  open dataset for autonomous driving and its application,'' \emph{IEEE
  Transactions on Pattern Analysis and Machine Intelligence}, vol.~42, no.~10,
  pp. 2702--2719, 2019.

\bibitem{c4}
H.~Caesar, V.~Bankiti, A.~H. Lang, S.~Vora, V.~E. Liong, Q.~Xu, A.~Krishnan,
  Y.~Pan, G.~Baldan, and O.~Beijbom, ``{nuScenes}: A multimodal dataset for
  autonomous driving,'' in \emph{Proceedings of the IEEE/CVF Conference on
  Computer Vision and Pattern Recognition}, 2020, pp. 11\,621--11\,631.

\bibitem{c5}
P.~Sun, H.~Kretzschmar, X.~Dotiwalla, A.~Chouard, V.~Patnaik, P.~Tsui, J.~Guo,
  Y.~Zhou, Y.~Chai, B.~Caine \emph{et~al.}, ``Scalability in perception for
  autonomous driving: Waymo open dataset,'' in \emph{Proceedings of the
  IEEE/CVF Conference on Computer Vision and Pattern Recognition}, 2020, pp.
  2446--2454.

\bibitem{c6}
P.~Xiao, Z.~Shao, S.~Hao, Z.~Zhang, X.~Chai, J.~Jiao, Z.~Li, J.~Wu, K.~Sun,
  K.~Jiang \emph{et~al.}, ``Pandaset: Advanced sensor suite dataset for
  autonomous driving,'' in \emph{2021 IEEE International Intelligent
  Transportation Systems Conference (ITSC)}.\hskip 1em plus 0.5em minus
  0.4em\relax IEEE, 2021, pp. 3095--3101.

\bibitem{c7}
Y.~Wang, X.~Chen, Y.~You, L.~E. Li, B.~Hariharan, M.~Campbell, K.~Q.
  Weinberger, and W.-L. Chao, ``Train in {Germany}, test in the {USA}: Making
  {3D} object detectors generalize,'' in \emph{Proceedings of the IEEE/CVF
  Conference on Computer Vision and Pattern Recognition}, 2020, pp.
  11\,713--11\,723.

\bibitem{c8}
J.~Yang, S.~Shi, Z.~Wang, H.~Li, and X.~Qi, ``St3d: Self-training for
  unsupervised domain adaptation on 3d object detection,'' in \emph{Proceedings
  of the IEEE/CVF Conference on Computer Vision and Pattern Recognition}, 2021,
  pp. 10\,368--10\,378.

\bibitem{c9}
D.~Tzutalin, ``Labelimg,'' \emph{GitHub Repository}, vol.~6, 2015.

\bibitem{c10}
B.~C. Russell, A.~Torralba, K.~P. Murphy, and W.~T. Freeman, ``Labelme: a
  database and web-based tool for image annotation,'' \emph{International
  Journal of Computer Vision}, vol.~77, no.~1, pp. 157--173, 2008.

\bibitem{c11}
E.~Li, S.~Wang, C.~Li, D.~Li, X.~Wu, and Q.~Hao, ``Sustech points: A portable
  3d point cloud interactive annotation platform system,'' in \emph{2020 IEEE
  Intelligent Vehicles Symposium (IV)}.\hskip 1em plus 0.5em minus 0.4em\relax
  IEEE, 2020, pp. 1108--1115.

\bibitem{c12}
H.~A. Arief, M.~Arief, G.~Zhang, Z.~Liu, M.~Bhat, U.~G. Indahl, H.~Tveite, and
  D.~Zhao, ``Sane: smart annotation and evaluation tools for point cloud
  data,'' \emph{IEEE Access}, vol.~8, pp. 131\,848--131\,858, 2020.

\bibitem{c13}
T.~Wang, C.~He, Z.~Wang, J.~Shi, and D.~Lin, ``Flava: Find, localize, adjust
  and verify to annotate lidar-based point clouds,'' in \emph{Adjunct
  Publication of the 33rd Annual ACM Symposium on User Interface Software and
  Technology}, 2020, pp. 31--33.

\bibitem{c14}
W.~Zimmer, A.~Rangesh, and M.~Trivedi, ``3d bat: A semi-automatic, web-based 3d
  annotation toolbox for full-surround, multi-modal data streams,'' in
  \emph{2019 IEEE Intelligent Vehicles Symposium (IV)}.\hskip 1em plus 0.5em
  minus 0.4em\relax IEEE, 2019, pp. 1816--1821.

\bibitem{c15}
B.~Wang, V.~Wu, B.~Wu, and K.~Keutzer, ``Latte: accelerating lidar point cloud
  annotation via sensor fusion, one-click annotation, and tracking,'' in
  \emph{2019 IEEE Intelligent Transportation Systems Conference (ITSC)}.\hskip
  1em plus 0.5em minus 0.4em\relax IEEE, 2019, pp. 265--272.

\bibitem{c16}
F.~Wirth, J.~Quehl, J.~Ota, and C.~Stiller, ``Pointatme: efficient 3d point
  cloud labeling in virtual reality,'' in \emph{2019 IEEE Intelligent Vehicles
  Symposium (IV)}.\hskip 1em plus 0.5em minus 0.4em\relax IEEE, 2019, pp.
  1693--1698.

\bibitem{c17}
C.~Sager, P.~Zschech, and N.~K{\"u}hl, ``labelcloud: A lightweight labeling
  tool for domain-agnostic 3d object detection in point clouds,''
  \emph{Computer-Aided Design and Applications}, vol.~19, no.~6, p. 1191, 2022.

\bibitem{c19}
M.~Ester, H.-P. Kriegel, J.~Sander, X.~Xu \emph{et~al.}, ``A density-based
  algorithm for discovering clusters in large spatial databases with noise.''
  in \emph{kdd}, vol.~96, no.~34, 1996, pp. 226--231.

\bibitem{c23}
R.~E. Kalman, ``A new approach to linear filtering and prediction problems,''
  1960.

\bibitem{c21}
H.~W. Kuhn, ``The hungarian method for the assignment problem,'' \emph{Naval
  research logistics quarterly}, vol.~2, no. 1-2, pp. 83--97, 1955.

\bibitem{c1}
J.~Deng, S.~Shi, P.~Li, W.~Zhou, Y.~Zhang, and H.~Li, ``Voxel r-cnn: Towards
  high performance voxel-based 3d object detection,'' in \emph{Proceedings of
  the AAAI Conference on Artificial Intelligence}, vol.~35, no.~2, 2021, pp.
  1201--1209.

\bibitem{c20}
C.-Y. Wang, A.~Bochkovskiy, and H.-Y.~M. Liao, ``Yolov7: Trainable
  bag-of-freebies sets new state-of-the-art for real-time object detectors,''
  \emph{arXiv preprint arXiv:2207.02696}, 2022.

\bibitem{c22}
N.~Ravi, J.~Reizenstein, D.~Novotny, T.~Gordon, W.-Y. Lo, J.~Johnson, and
  G.~Gkioxari, ``Accelerating 3d deep learning with {PyTorch3D},'' \emph{arXiv
  preprint arXiv:2007.08501}, 2020.

\end{thebibliography}

\end{document}